\pdfoutput=1
\documentclass[11pt]{article}

\usepackage[]{acl}

\usepackage{times}
\usepackage{latexsym}

\usepackage{amsmath}
\usepackage{multirow}
\usepackage{tabularx}
\usepackage{adjustbox}

\usepackage[many]{tcolorbox}    	% for COLORED BOXES (tikz and xcolor included)
\usepackage{setspace}               % for LINE SPACING
\usepackage{multicol}               % for MULTICOLUMNS

\usepackage[T1]{fontenc}
\usepackage[utf8]{inputenc}

\usepackage{microtype}

\usepackage{inconsolata}

\usepackage{colortbl}
\usepackage{multirow}
\usepackage{graphicx}

\definecolor{main}{HTML}{5989cf}    % setting main color to be used
\definecolor{sub}{HTML}{cde4ff}     % setting sub color to be used
\newtcolorbox{boxD}{
    colback = sub, 
    colframe = main, 
    boxrule = 0pt, 
    toprule = 3pt, % top rule weight
    bottomrule = 3pt, % bottom rule weight
    breakable
}

\title{\textsc{TaxoAlign}: Scholarly Taxonomy Generation Using Language Models}

\author{Avishek Lahiri\textsuperscript{*} \hfill Yufang Hou\textsuperscript{$\dag$}  \hfill Debarshi Kumar Sanyal\textsuperscript{*} \\
  \textsuperscript{*}Indian Association for the Cultivation of Science, Kolkata, India \\
  \textsuperscript{$\dag$}IT:U Interdisciplinary Transformation University Austria, Linz, Austria \\
  \texttt{avisheklahiri2014@gmail.com},\\ \texttt{yufang.hou@it-u.at}, \texttt{debarshi.sanyal@iacs.res.in} \\}

\begin{document}
\maketitle
\begin{abstract}

Taxonomies play a crucial role in helping researchers structure and navigate knowledge in a hierarchical manner. They also form an important part in the creation of comprehensive literature surveys. The existing approaches to automatic survey generation do not compare the structure of the generated surveys with those written by human experts. To address this gap, we present our own method for automated taxonomy creation that can bridge the gap between human-generated and automatically-created  taxonomies. For this purpose, we create the \textsc{CS-TaxoBench} benchmark which consists of $460$ taxonomies that have been extracted from human-written survey papers. We also include an additional test set of $80$ taxonomies curated from conference survey papers. We propose \textsc{TaxoAlign}, a three-phase topic-based instruction-guided method for scholarly taxonomy generation. Additionally, we propose a stringent automated evaluation framework that measures the structural alignment and semantic coherence of automatically generated taxonomies in comparison to those created by human experts. We evaluate our method and various baselines on \textsc{CS-TaxoBench}, using both automated evaluation metrics and human evaluation studies. The results show that \textsc{TaxoAlign} consistently surpasses the baselines on nearly all metrics. The code and data can be found at \url{https://github.com/AvishekLahiri/TaxoAlign}.
%Taxonomies are important tools for researchers for the hierarchical organization of knowledge.
%The automated creation of taxonomies tests the long-context reasoning capabilities of large language models (LLMs) as well as their ability to generate abstractive and precise information from vast amounts of data. 
%None of the existing approaches to automatic survey generation compare the structure of the generated surveys with those written by human experts. 
%Additionally, we present a stringent automated evaluation method for comparing taxonomies that %tests 
%assesses 
%the alignment of taxonomy trees as well as their semantic coherence between the human created ones and automatically generated ones. % 
%We evaluate our approach and various baselines using our benchmark and automated evaluation method along with human evaluation. 

\end{abstract}

\section{Introduction}

\begin{figure}[h]
  \centering
  \includegraphics[width=\linewidth]{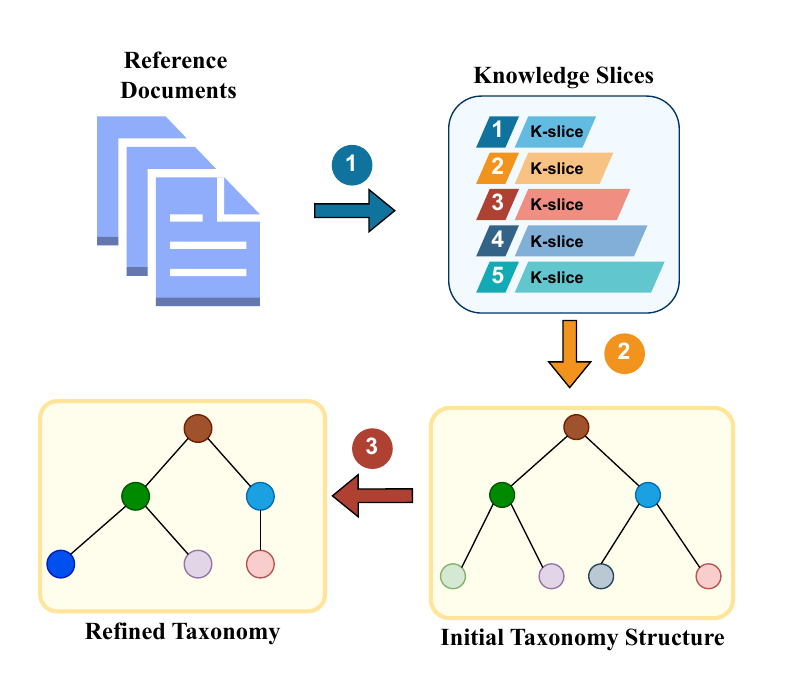}
  \caption{Schematic representation of \textsc{TaxoAlign}: (1) Knowledge Slice Creation (2) Taxonomy Verbalization (3) Taxonomy Refinement}
  \label{fig:TaxoAlign-framework}
\end{figure}

%In scientific research, a taxonomy is constructed on a topic which has a well-defined scope such that the research results on that topic may be integrated under a single hood such that a better understanding of the topic gets developed among interested researchers and industry practitioners alike. 
In scientific research, a taxonomy is constructed around a well-defined topic to integrate relevant research findings under a unified framework, thereby facilitating deeper understanding among researchers and industry practitioners alike.
In the general domain, taxonomies have been proven to be useful tools \cite{wang-etal-2017-short}, which exhibit the capability of enhancing the performance of various Natural Language Processing (NLP) and Information Retrieval (IR) tasks such as question answering \cite{HARABAGIU_MAIORANO_PAŞCA_2003, Yang_Zou_Wang_Yan_Wen_2017}, textual entailment \cite{geffet-dagan-2005-distributional}, personalized recommendation \cite{Zhang_personalized_recommendation}, query understanding \cite{hua_query_understanding},  information extraction \cite{hou-etal-2019-identification,sahinuc-etal-2024-efficient} and knowledge graph construction \cite{hou-etal-2021-tdmsci, mondal-etal-2021-end}. Taxonomies have also found a place in real-world deployment applications such as biomedical systems \cite{10.1093/nar/gkt1026}, information management \cite{Nickerson_inf_management} and e-commerce \cite{Aanen_e_commerce, Octet}. 

\begin{figure*}[h]
  \centering
  \includegraphics[width=0.9\linewidth]{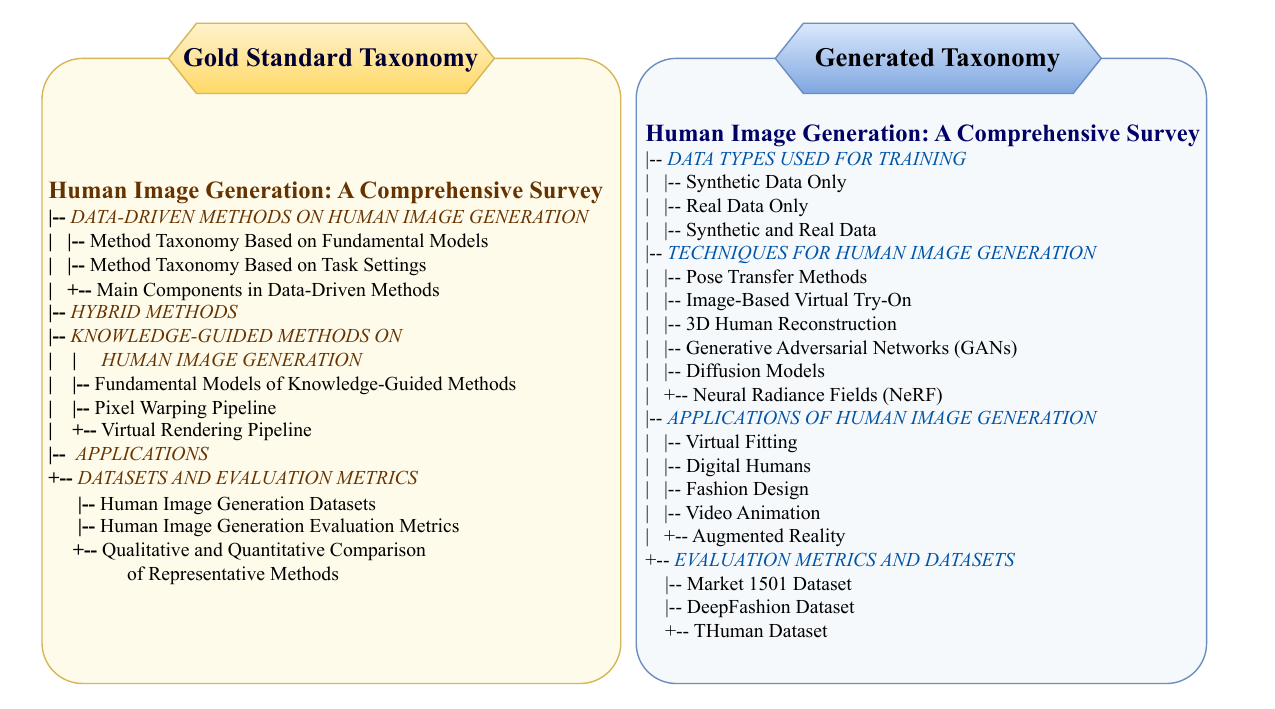}
  %\textcolor{red}{(YH: please make the curve of the box sharper}\\
  \caption{A comparison of a gold standard taxonomy tree and a generated taxonomy tree using \textsc{TaxoAlign} for the topic "Human Image Generation: A Comprehensive Survey". The generated taxonomy shown here uses Mistral-7B-Instruct-v0.3 for creation of knowledge slices and Llama-3.1-T\"{u}lu-3-8B for the taxonomy verbalization component and GPT-4o-mini for refining the generated taxonomy.}
  \label{fig:Taxonomy-gen-example}
\end{figure*}

In this paper, we conceptualize the task of \emph{automated scholarly taxonomy generation}. When given the taxonomy topic and a set of related reference papers, the task for the model is to reason over this large set of reference documents and generate a taxonomy tree that is both concise and provides maximal coverage of the reference documents. 
Automating this process reduces the time, effort, and energy researchers spend organizing research within a specific topic.
%Automating this process helps to reduce the effort, energy and time that researchers put in to effectively organize research under a certain topic. 
%It also helps researchers to navigate quickly to a sub-topic of their interest in the taxonomy.

%Large language models (LLMs) are used all over the world today for their generative capabilities including Machine Comprehension or the Natural Language Understanding. Similarly, language models can also be applied to generate taxonomies.
% Although large language models (LLMs) have been used in a wide range of tasks nowadays, 
% %the automated creation of taxonomies tests the long-context reasoning capabilities of large language models (LLMs) as well as their ability to generate abstractive and precise information from vast amounts of data.
% %However, language models 
% they face constraints in the comprehension and execution of domain-specific tasks \cite{li2024scilitllmadaptllmsscientific, cai-etal-2025-sciassess}. Moreover, there are some limitations regarding their reasoning capability over long-contexts due to context-window limitations \cite{liu-etal-2024-lost}. 
While Large Language Models (LLMs) are increasingly applied across a wide range of tasks, they face notable challenges in understanding and performing domain-specific tasks \cite{li2024scilitllmadaptllmsscientific, cai-etal-2025-sciassess}. Additionally, their reasoning over long contexts is limited by constraints in context window size \cite{liu-etal-2024-lost}.
In our initial pilot study,  
%for this work, 
we observed that prompting LLMs could generate some topic-related sub-topics, but they were not remotely aligned to the human-written taxonomy. For example, ``\emph{3D Models and Mapping}'', ``\emph{Generative Adversarial Network}'' and ``\emph{Data and Annotation}'' were the first-level nodes/sub-topics that were generated by T\"{u}lu3 \cite{lambert2025tulu3pushingfrontiers} for the topic ``\emph{Human Image Generation: A Comprehensive Survey}'' when supplied with all the summaries of the cited documents.  %Modified: \textcolor{red}{(DKS: What is a ref. summary?)}. %Therefore, 
In general, 
the generated nodes were distantly relevant to the topic but were not at close to the human-written ones as shown in Figure \ref{fig:Taxonomy-gen-example} (left).
%Modified: \textcolor{red}{(YH: it should be fig 2 (left))}.

Recent prior work has focused extensively on the end-to-end automated creation of survey papers \cite{AutoSurvey, liang2025surveyxacademicsurveyautomation, yan2025surveyforgeoutlineheuristicsmemorydriven, kang2025researcharenabenchmarkinglargelanguage}. %In contrast, scholarly taxonomy generation is a relatively untrodden path with no open-source well-articulated data resource having been developed for this task. None of the prior work compares the structure of the generated surveys with
%those written by human experts. There is also a lack of an evaluation suite that can assess both the structural similarity and semantic coherence between the generated and human-written taxonomy trees.
In contrast, scholarly taxonomy generation remains a relatively unexplored area, with no well-articulated open-source data resources currently available for this task. Moreover, prior work does not compare the structure of generated surveys with those authored by human experts. There is also a lack of an evaluation framework capable of assessing both the structural similarity and semantic coherence between automatically generated and human-written taxonomy trees.

%We list or main contributions as follows,

To address this gap, we curate and release \textsc{CS-TaxoBench}, a comprehensive benchmark that is designed for the task of scholarly taxonomy generation (Section \ref{sec:dataset}). Our benchmark consists of $460$ human-written taxonomies (accompanied by their corresponding reference papers) that have been extracted from survey articles published in Computer Science journals in $2020-2024$.
%Modified: \textcolor{red}{(DKS: rephrase? survey articles published in computer science journals in $2020-2024$)}. 
We also curate an additional test set made up of 80 taxonomies extracted from conference survey papers.
 
We further develop \textsc{TaxoAlign}, our own intuitive LLM-based pipeline for generating taxonomies (Section \ref{sec:method}).
%which leverages the use of only such information that is relevant to the taxonomy topic to generate the taxonomy tree
\textsc{TaxoAlign} consists of three parts: \emph{Knowledge Slice Creation}, \emph{Taxonomy Verbalization} and \emph{Taxonomy Refinement}. We compare our method with a range of baselines to show the effectiveness of our method. Figure \ref{fig:TaxoAlign-framework} demonstrates our proposed framework, while Figure \ref{fig:Taxonomy-gen-example} (right) 
%Modified: \textcolor{red}{(YH: it sould be Fig 2 (right))} 
shows an example of a taxonomy generated using \textsc{TaxoAlign}.

Finally, we present an automated evaluation framework for the comparison of taxonomy-tree structures (Section \ref{sec:evaluation}). For this purpose, we develop two metrics of our own -- the average degree score metric to judge structural similarity and the level-order traversal comparison metric to judge semantic similarity. %Apart from this, we use the soft recall and entity recall metrics (that were originally proposed to evaluate outline-generation). We also use LLM-as-a-judge to judge the generated taxonomies. \textsc{TaxoAlign} outperforms all the baselines on almost all the metrics including human evaluation.
In addition, we adopt soft recall and entity recall metrics, originally proposed for evaluating outline generation, to assess the quality of the generated taxonomies \cite{FRANTI2023115, shao-etal-2024-assisting, kang2025researcharenabenchmarkinglargelanguage}. %\textcolor{red}{(DKS: citation?)} 
We also employ LLM-as-a-judge for qualitative evaluation. \textsc{TaxoAlign} outperforms all baselines across nearly all metrics, including human evaluation. To facilitate future research, we make our code and dataset publicly available at \url{https://github.com/AvishekLahiri/TaxoAlign}.

%\textsc{TaxoAlign} makes a significant contribution to the field of automatic scholarly taxonomy generation by enabling systematic comparison of existing approaches. 
%To support and encourage future research, we will release the code and dataset upon publication.

\section{Related Work}
\paragraph{Taxonomy Construction.}
Taxonomy learning has been attempted in NLP through the decades by captalizing on the semantic relations in text \cite{hearst-1992-automatic, pantel-pennacchiotti-2006-espresso, Suchanek_et_al, PONZETTO20111737, RIOSALVARADO20135907, TaxoLearn, Liu_Automatic_taxonomy, semantic_GrowBag, 4912206, 7236916, kozareva-hovy-2010-semi, velardi-etal-2013-ontolearn}. 
Recent approaches such as HiGTL \cite{hu2025taxonomytreegenerationcitation} and the method of \newcite{martel_taxonomy} introduce graph- and clustering-based techniques for taxonomy learning. Most of these methods use pattern-based or clustering-based methods, whereas \textsc{TaxoAlign} leverages the power of LLMs to construct taxonomies in the scientific domain.
%HiGTL \cite{hu2025taxonomytreegenerationcitation}  and \cite{martel_taxonomy} are more recent works that propose graph and clustering-based methods for taxonomy learning.

\paragraph{Scientific Survey Generation and Knowledge Synthesis.}
Recently, there has been some interest among researchers to generate surveys from a corpus of research papers. AutoSurvey \cite{AutoSurvey}, \textsc{SurveyX} \cite{liang2025surveyxacademicsurveyautomation}, ResearchArena \cite{kang2025researcharenabenchmarkinglargelanguage}, Qwen-long \cite{lai2024instructlargelanguagemodels} and \textsc{SurveyForge} \cite{yan2025surveyforgeoutlineheuristicsmemorydriven} are some of the prominent techniques proposed for this task.
In literature-based knowledge synthesis, LLMs have been used to generate scientific leaderboards \cite{sahinuc-etal-2024-efficient,timmer2025positionpaperautomaticgeneration},  literature review tables \cite{newman-etal-2024-arxivdigestables}, or to synthesize biomedical evidence in the format of forest plots \cite{pronesti-etal-2025-query, pronesti2025enhancingstudylevelinferenceclinical}. Our work
focuses on scientific survey taxonomy generation and contributes to the broader agenda
of AI for Science \cite{eger2025transformingsciencelargelanguage}.

\section{\textsc{CS-TaxoBench}}
\label{sec:dataset}

\begin{figure*}
    \centering
    \includegraphics[width=0.9\linewidth]{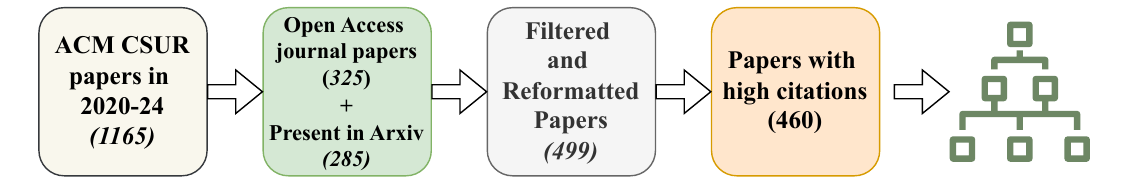}
    \caption{An overview of the pipeline for the curation of our dataset.}
    \label{fig:data-pipeline}
\end{figure*}

\subsection{Overview}

In graph theory, a tree is defined as an undirected connected graph with no cycles. A taxonomy tree $T$ is a tree in which the root represents the taxonomy topic and the child nodes represent sub-topics and grandchild nodes represent more fine-grained topics.
%Scholarly Taxonomy Generation is a challenging task in the domain of scientific literature understanding which necessitates multi-document reasoning over scientific research papers. 
Survey papers typically propose a taxonomy which is expanded upon in the sections and sub-sections of the paper. Therefore, in most cases, the structure of the paper closely mirrors the nodes and connections in the taxonomy tree.
%Survey papers typically begin by presenting a taxonomy in either tabular or graphical form. The rest of the paper then provides a comprehensive study of the subtopics represented in this taxonomy. In most cases, the structure of the paper closely mirrors the nodes and connections in the taxonomy tree.
%The general trend in survey papers is to propose a taxonomy at the beginning of the paper in a tabular or graphical form. The remainder of the paper makes a comprehensive study of the sub-topics captured by the taxonomy. Usually, the section structure of the paper bears a strong correspondence with the nodes and their connectivity in the taxonomy tree. 
%with appropriate  and then expand each part and sub-part of the taxonomy is expanded through the paper in the form of sections and sub-sections. 
%We leverage this pattern 
%to our benefit 
%for the extraction of 
%to extract scholarly taxonomies from survey papers. Our taxonomy extraction module extracts outlines from survey papers, and analyzes them to arrive at the final taxonomy.
We leverage this pattern to extract scholarly taxonomies, using our taxonomy extraction module to first derive outlines from survey papers and then analyze them to construct the final taxonomy.

\subsection{Desiderata}

We list out the desiderata we used for selecting taxonomies for inclusion in our benchmark dataset. Our overall goal was to ensure that the selected taxonomies are of high-quality with a logical flow and each node is grounded in a set of reference papers.
We decide on the following desiderata for curating our dataset: 
(1) each taxonomy should be based on a specific research topic, and the taxonomy should provide optimal coverage of the given topic; 
(2) the taxonomies should be human-made, i.e., they should not be generated artificially; 
(3) the taxonomy trees should be multi-layered, i.e., they should have at least two levels. 
%Modified: \textcolor{red}{(DKS: layers or levels?)}

\subsection{Automated Taxonomy Extraction}

\textbf{Data Source Selection}:  Survey papers form a rich resource for taxonomies in scientific literature. Therefore, we select survey papers as our primary source of data. To mitigate the risk of data contamination while ensuring that the 
%survey 
papers in consideration are of high quality, we select survey papers from ``\textit{ACM Computing Surveys}'' (ACM CSUR), which is a highly reputed journal in the field of Computer Science research with an Impact Factor of $23.8$.\footnote{\url{https://dl.acm.org/journal/csur}} This is one of the top venues in Computer Science that publishes survey papers relating to the areas of computing research and practice.

\textbf{Research Paper Selection}: We select a time frame of five years between $2020$ to $2024$ for the purpose of the creation of our dataset. %There are a total of $1,165$ papers that have been accepted in the ACM CSUR journal during this time, among which $325$ are open-access articles and $285$ articles have their copies available in ArXiv\footnote{\url{https://arxiv.org/}}. We only select the open-access articles and articles from Arxiv due to licensing restrictions.
A total of 1,165 papers were accepted in ACM CSUR during this period, of which 325 are open-access and 285 have copies available on arXiv.\footnote{\url{https://arxiv.org/}}
 Due to licensing restrictions, we include only the open-access and arXiv articles in our study.

%\textbf{Filtering}: We use Docling \cite{Docling} to extract the text from the research paper PDFs. Since the papers from ArXiv are not always in the same format, we remove those with %separate 
%noisy format and Docling parsing errors. %Therefore, 
%After these stages of filtering, we are left with $499$ papers. 
%For example, we do not take into consideration articles for which only a small number of reference papers have been retrieved from the text.
\textbf{Filtering}: We use Docling \cite{Docling} to extract text from the research paper PDFs. Since arXiv papers are not always in a consistent format, we remove those with noisy layouts or Docling parsing errors. After filtering, 499 papers remain.

\textbf{Reference Paper Matching}: We retrieve the abstracts of the reference papers by parsing data from Semantic Scholar\footnote{\url{https://www.semanticscholar.org/me/research}}, which hosts approximately 214 million research documents.
If fewer than 50\% of a survey paper’s references are available %with abstracts 
on Semantic Scholar, we exclude that paper from the final version of our proposed datasets. Following this criterion, 39 out of the original 499 papers are excluded, leaving us with a final set of 460 papers. The average percentage of available citations in the final set of papers is shown in Table \ref{tab:stats}.

\textbf{Taxonomy Extraction}: We extract the headings, subheadings, and sub-subheadings from the retrieved text of the survey paper to construct the taxonomy. The title of the survey paper is treated as the overall taxonomy topic. We discard all headings that contain terms such as ``Introduction'', ``related work'', ``problem formulation'',  ``summary'', ``conclusion'', ``result'', ``future'', ``discussion'', ``background'' and ``overview'' as they do not contribute to the core taxonomy. We also remove those nodes for which we cannot extract reference papers from Semantic Scholar. 
%This is done to ensure that the final extracted taxonomy contains only those labels that are highly pertinent to the survey topic.

\begin{table}[h]
    \centering
    \begin{tabular}{lc}
    \hline
        \textbf{Statistics}                     &  \textbf{Value} \\ \hline
        No. of taxonomies in train set          &  $400$ \\
        No. of taxonomies in test set          &   $60$ \\
        Total no. of cited papers               &  $79,027$\\
        Cited papers present in S2              &  $60,373$\\
        \% of available cited papers    &  $76.40 \%$ \\ \hline
        
    \end{tabular}
    \caption{Statistics of our proposed benchmark. Note that here S2 refers to Semantic Scholar \cite{lo-etal-2020-s2orc}.}
    \label{tab:stats}
\end{table}

\subsection{Dataset Statistics}

Our entire benchmark contains $460$ taxonomies along with the reference papers that are used to build each taxonomy. The details about the statistics of our benchmark are present in Table \ref{tab:stats}. %It is seen that 
On average, there are around $131$ reference papers for each taxonomy in our benchmark.

\subsection{Manual Annotation}
%Benchmark Construction}

To evaluate the taxonomy extraction method, we manually annotate a set of $10$ taxonomy trees from as many survey papers. We use a Python package, TreeLib\footnote{\url{https://treelib.readthedocs.io/en/latest/}}, to annotate these taxonomies from their respective survey papers. The papers were selected such that there were explicitly-defined taxonomies in them. To compare the annotated and generated trees, we compare the paths from each node to the root node. Each path is treated as an individual element. The precision, recall and F1 between the annotated and extracted taxonomies were found to be $83.92\%$, $94.35\%$ and $88.83\%$ respectively, thereby demonstrating a high degree of correlation. The only errors originated due to some general section headers like ``Two sea changes in Natural Language Processing'' \cite{prompt_survey}, which were not present in the annotated trees.

\begin{figure*}
    \centering
    \includegraphics[width=\linewidth]{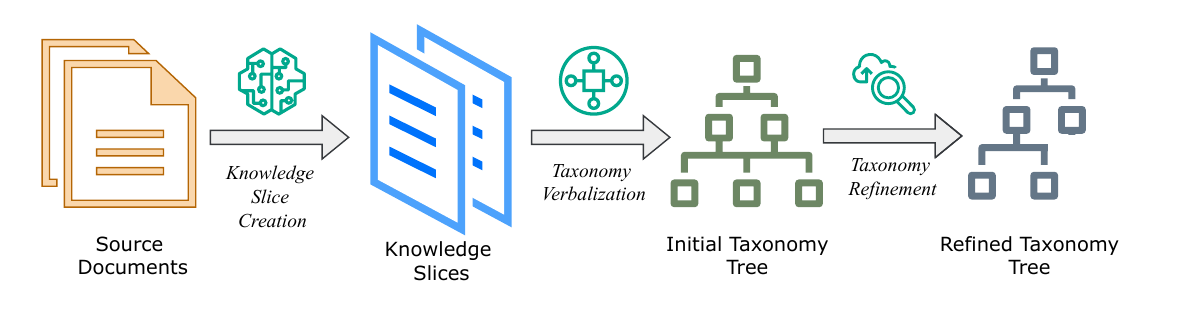}
    \caption{An overview of the proposed \textsc{TaxoAlign} pipeline.}
    \label{fig:method}
\end{figure*}

%\subsection{Conference Survey Papers Test Set}
\subsection{Additional Test Set: Survey Papers in Conferences}
\label{conf_papers_section}

We create an additional new test set from conference survey papers for testing on a different distribution of survey papers. For this purpose, we chose survey papers published in 2024 in IJCAI (which has a dedicated survey track) and all the ACL* conferences (ACL, NAACL, EMNLP and EACL). There are a total of 86 survey papers in these conferences in 2024. We carry out our proposed filtering strategies, which narrows down the total number of papers to 80. Therefore, we have 80 taxonomy trees with an average of about 71 reference papers for each tree. This set is solely used as a test set.

\section{\textsc{TaxoAlign}}
\label{sec:method}

%\textcolor{red}{(DKS: Suddenly the acronym \textit{LLM} ceases to be used.}\\
%We formalize the task of taxonomy generation in scholarly documents in this section. 
\subsection{Task Formulation} 
%Given a corpus of documents $D$ and a taxonomy topic $t$, the objective is to construct a taxonomy tree $T$, such that $T$ is representative of all the documents in $D$.
Given a corpus of documents $D$ and a topic $t$, the task is to automatically construct a hierarchical taxonomy tree $T$ whose nodes represent relevant topics and subtopics derived from $D$. The goal for $T$ is to comprehensively and meaningfully categorize the information contained within the entire corpus that relates to topic $t$.

To solve this task, we propose a method \textsc{TaxoAlign}, comprising three components: Knowledge Slice Creation, Taxonomy Verbalization and Taxonomy Refinement. 
%Figure \ref{fig:method} shows an outline of the three stages of our proposed method.
Figure \ref{fig:method} shows the three-stage pipeline of our proposed method.

\subsection{Knowledge Slice Creation}

%In this step, we utilize a large language model to find the pieces of information within a research article that is highly relevant to the taxonomy topic. We refer to these pieces of information as ``knowledge slices''. Therefore, each knowledge slice is a highlighted information piece in the research document which also relates strongly to the taxonomy topic. 
In this step, we use a LLM to identify segments of text within each research article that are highly relevant to the taxonomy topic. We refer to these segments as knowledge slices. 

%This stage helps in extracting the information which can provide accurate guidance to the model in the future steps about the nature of topics and sub-topics that are present in the cited papers related to the taxonomy topic.
This stage helps in extracting information that guides the model in later steps about the topics and subtopics present in the cited papers related to the taxonomy.
More importantly, the number of cited papers in a survey paper is quite extensive, thereby making it quite impossible to fit all the papers in the model's input context window. This step ensures that all the cited papers can be accommodated within the context length of recent LLMs. %large language models.

\subsection{Taxonomy Verbalization}

We opt for the instruction tuning of 
%a generative large language model
an LLM 
with the most pertinent taxonomy topic-related information from the reference papers that has been extracted in the previous stage.

The main objective is to teach the model to generate meaningful and concise taxonomies which are grounded in the given information, and most importantly, teach the model to learn the structure of the taxonomy trees. 
Finetuning helps in preserving the structure of the taxonomy in a major way, which is a feature that is lacking in the direct prompting-based methods.  

\subsection{Taxonomy Refinement}

The verbalization phase is followed by a refinement stage, which evaluates and refines the connections between the parent and the child nodes. It checks whether each node is grounded in the document knowledge slices. 
%highlights. %\textcolor{red}{YH: you mean knowledge slices here?}\textcolor{blue}{Yes Ma'am}.
If the tree contains too few nodes, it expands the node set to achieve a greater coverage of the documents using their corresponding knowledge slices. This refinement strategy is executed by prompting an LLM with stronger reasoning capabilities than those used in the previous stages.

\section{Evaluation}
\label{sec:evaluation}
%We describe here the new evaluation approaches developed by us as well as the existing evaluation methods in this domain. The key challenge in comparing the generated tree with the gold tree lies in aligning the two trees.
We present both the new evaluation approaches we have developed and the existing methods used in this domain. The main challenge in comparing a generated tree with the gold tree lies in aligning the two structures. 
%The two trees should have similar structure in addition to being semantically similar. Aligning is tough because the trees may have significant differences in their hierarchical structures or low lexical overlap, 
The two trees should be structurally similar as well as semantically aligned. However, alignment is challenging because the trees may differ significantly in their hierarchical structures or exhibit low lexical overlap, 
%instead of being semantically similar, 
as has been encountered in similar problems such as table schema alignment evaluation \cite{newman-etal-2024-arxivdigestables}. 

\subsection{Average Degree Score}

The first condition for two trees to be considered similar is that their structures should be similar. We design this metric to judge the structural similarity of the generated tree and the gold standard tree.

In a graph, the average degree is calculated as the average number of edges connected to a node in the graph. For any tree with $N$ nodes, the number of edges is $N-1$, which gives the average degree of $2(N-1)/N$. Therefore, to judge the structural similarity between the gold standard tree $T$ and the predicted tree $T'$, we find the average degree score $\Delta$, which is the ratio between the average degree of $T'$ and the average degree of $T$. 

\begin{equation}
    \Delta = \dfrac{\sum_{i=1}^{m} d(T'_i)}{\sum_{i=1}^{n} d(T_i)}
\end{equation}
%\textcolor{red}{(DKS: In the eq., isn't it $d(t)$ (and $d(t')$)?}\\: Response: No Sir, it is because $T$ and $T'$ represent the gold and predicted trees respectively.
where, $d(t)$ represents the degree of a node $t$, and $m$ and $n$ are the number of nodes in the trees $T$ and $T'$ respectively. In the ideal case, the value of $\Delta$ should be $1$. If the generated tree $T'$ is more branched out than the original tree $T$, then the value of $\Delta$ is greater than $1$, while if $T'$ is less branched out than it should be, then the value of $\Delta$ is less than $1$. We report the final score as the mean of all the average degree scores in the test corpus.

\subsection{Level-order Traversal Comparison}

The hierarchical structure of a tree makes it difficult to implement standard text generation metrics such as BLEU \cite{bleu}, ROUGE \cite{lin-2004-rouge}, or BERTScore \cite{bert_score}. Therefore, we propose a metric to compare the gold standard tree and the generated tree using level-order traversal. 
%In level-order traversal, 
More specifically, 
we traverse the tree in such a way that all the nodes present in the same level are traversed completely before traversing the next level. After converting the entire tree into a single list through level-order traversal, we calculate the the corpus-level BLEU-2, ROUGE-L and  BERTScore.

\subsection{Node Soft Recall and Node Entity Recall}

%Following the evaluation protocol in prior work \cite{FRANTI2023115, shao-etal-2024-assisting, kang2025researcharenabenchmarkinglargelanguage}, the Node Soft Recall and the Node Entity Recall are defined as follows. These metrics compare the generated and the ground-truth taxonomy trees using semantic similarity and lexical overlap between them, respectively.
We use the metrics \emph{Node Soft Recall} and \emph{Node Entity Recall}, following the evaluation protocol in prior work \cite{FRANTI2023115, shao-etal-2024-assisting, kang2025researcharenabenchmarkinglargelanguage}.   These metrics compare the generated and the ground-truth taxonomy trees using semantic similarity and lexical overlap between them, respectively. 
%They are defined as follows.
Node Soft Recall (NSR) is dependent on soft cardinality of a tree \cite{soft_cardinality}, which is given by,

\begin{equation} \label{eq:2}
    c(T) = \sum_{i=1}^{n} \dfrac{1}{n\sum_{j=1}^{n} \operatorname{Sim}(T_i, T_j)}
\end{equation}
%Sim(T_i, T_j) = cos(embed(T_i), embed(T_j))
where $\operatorname{Sim}(T_i, T_j)$ is the cosine similarity between the \textsc{Sentence-BERT}  \cite{reimers2019sentence} embeddings of the taxonomy trees $T_i$ and $T_j$. The Node Soft Recall between two trees $T$ and $T'$ is defined as,

\begin{equation}
    \text{NSR}(T, T') = \dfrac{c(T) + c(T') - c(T \cup T')}{c(T')}
\end{equation}

Since in most baselines, there is a large mismatch between the number of nodes of the generated taxonomy tree and the original tree, we tweak the original heading soft recall in \newcite{shao-etal-2024-assisting} by inserting a normalizing factor in Eq. \ref{eq:2}, i.e., we divide by the number of nodes to offset the effect of a large node count in the tree.

Node Entity Recall (NER) between the gold standard tree $T$ and the generated tree $T'$ is defined as the percentage of entities that are present both in $T$ and $T'$. Formally, it can be expressed as,

\begin{equation}
    \text{NER}(T, T') = \dfrac{|\text{Ent}(T) \cap \text{Ent}(T')|}{|\text{Ent}(T)|}
\end{equation}

where $|{\text{Ent}}(T)|$ represents the number of entities in $T$. In our case, we track Noun Phrases (NP) for better coverage. We use the chunking model from FLAIR \cite{akbik-etal-2019-flair} for this purpose.

\subsection{LLM-as-a-Judge}

We prompt a stronger LLM, GPT-4.1, with the gold-standard taxonomy tree and the generated taxonomy tree, and ask it to evaluate the generated tree on a scale of $1$ to $5$ based on the structural similarity and the semantic similarity of the two trees. The LLM judges whether the generated tree aligns with the gold tree and whether they are coherent. The prompt is described in Appendix \ref{appendix:llm-eval}.

\section{Baselines}

In this section, we present the baselines against which we evaluate our method \textsc{TaxoAlign}.

\textbf{AutoSurvey} \cite{AutoSurvey}: The outline generation part of this method randomly divides the reference papers into several groups, which results in the creation of multiple outlines. The language model then amalgamates these outlines to construct a single comprehensive outline. %For fair evaluation with our method, we provide the reference papers unlike what was done in the original paper.
For a fair comparison with our method, we provide the reference papers, in contrast to the original work. 
We run only the outline generation step of AutoSurvey, instead of generating the whole article.

\textbf{STORM} \cite{shao-etal-2024-assisting}: The pre-writing stage of this method involves researching the given topic through simulated conversations. A draft outline is generated initially by prompting the LLM with the topic only. To generate the final outline, the LLM is prompted with the topic, simulated conversations as well as the draft outline. Like in AutoSurvey, here also we provide the pipeline with the reference papers.

\textbf{Topic only}: 
%In this baseline, we simply prompt the language model with the topic of that taxonomy and ask it to generate a corresponding tree that best suits the topic. This is mainly done to test how good the model can generate the tree based on only its parametric knowledge. 
In this baseline, we simply prompt a LLM with the taxonomy topic and ask it to generate a corresponding tree that best fits the topic. The primary goal is to evaluate how effectively the model can generate such a tree using only its parametric knowledge.

\textbf{Topic + Keyphrases}: We use a LLM to extract the top keyphrases from each of the reference (cited) papers that form the basis of the taxonomy. This provides the top phrases of each paper, enabling us to fit the essential content of all references within the limited context window of LLMs.
%This is another way in which we can get the highlights of all the reference paper documents to fit within the limited context window of LLMs. 
We then prompt another LLM to generate the taxonomy tree based on the set of keyphrases.

\textbf{\textsc{TaxoAlign} w/o Taxonomy Verbalization w/o Taxonomy Refinement}: We use the knowledge slices used in our proposed method. We simply prompt the model to generate the taxonomy tree based on these slices. This allows us to isolate and assess the effect of the latter stages of our own method, specifically those that occur after knowledge-slice creation. 
%i.e., following knowledge slice creation.

\begin{table*}[h]
    \centering
\begin{adjustbox}{width=\linewidth}
\begin{tabular}{c|c|c|ccc|c|c|c}
\hline
\multirow{2}{*}{\textbf{Method}}          & \multirow{2}{*}{\textbf{Model}}   & \multirow{2}{*}{\textbf{$\Delta$}} & \multicolumn{3}{c|}{\textbf{Level-order Traversal}}  & \multirow{2}{*}{\textbf{NSR}} & \multirow{2}{*}{\textbf{NER}} & \multirow{2}{*}{\textbf{\begin{tabular}[c]{@{}c@{}}LLM\\ judge\end{tabular}}} \\ \cline{4-6} & & & \multicolumn{1}{c|}{\textbf{BLEU-2}} & \multicolumn{1}{c|}{\textbf{ROUGE-L}} & \textbf{BERTScore} &        &        &                 \\ \hline

AutoSurvey & Prompt: GPT-4o-mini & 4.4659 & \multicolumn{1}{c|}{0.0016} & \multicolumn{1}{c|}{0.1784} & 0.8256 & 1.0903 & 0.1982 & 2.4333 \\ \hline

STORM & Prompt: GPT-4o-mini & 6.151 & \multicolumn{1}{c|}{0.0012} & \multicolumn{1}{c|}{0.1349} & 0.8166 & 1.0727 & 0.1539 & 2.2000 \\ \hline

Topic only & Prompt: T\"{u}lu & \textbf{1.4274} & \multicolumn{1}{c|}{0.0052} & \multicolumn{1}{c|}{0.2359} & 0.8376 & 1.4187 & 0.1373 & 2.0833 \\ \hline

\multirow{2}{*}{\begin{tabular}[c]{@{}c@{}}Topic\\  + \\ Keyphrases\end{tabular}}                     & \begin{tabular}[c]{@{}c@{}}Keyphrase: LLaMa; \\ Prompt: T\"                     {u}lu\end{tabular} & 4.4517 & \multicolumn{1}{c|}{0.0018}                           & \multicolumn{1}{c|}{0.1584} & 0.8134 & 1.1103 & 0.1491 &                         2.4167 \\ \cline{2-9} 
                   %& \begin{tabular}[c]{@{}c@{}}Keyphrase: LLaMa; \\ Prompt: SciLitLLM\end{tabular}          & 8.0766                  & \multicolumn{1}{c|}{0.0022}          & \multicolumn{1}{c|}{0.192}            & 0.8168              & 1.2170 & 0.1578 & 1.6833          \\ \cline{2-9} 
                   & \begin{tabular}[c]{@{}c@{}}Keyphrase: Mistral; \\ Prompt: T\"{u}lu\end{tabular}             & 4.91                    & \multicolumn{1}{c|}{0.0014}          & \multicolumn{1}{c|}{0.1432}           & 0.8100              & 1.0996 & 0.1640 & 2.4167          \\ \hline 
                   %& \begin{tabular}[c]{@{}c@{}}Keyphrase: Mistral; \\ Prompt: SciLitLLM\end{tabular}        & 6.6771                  & \multicolumn{1}{c|}{0.0029}          & \multicolumn{1}{c|}{0.1676}           & 0.8084              & 1.2522 & 0.1670 & 1.6500          \\ \hline
\multirow{6}{*}{\begin{tabular}[c]{@{}c@{}}\textsc{TaxoAlign}\\  w/o Tax. Verbaliz.\\ w/o Tax. Refine. \end{tabular}}                                           & \begin{tabular}[c]{@{}c@{}}K-Slice: LLaMa; \\ Prompt: T\"                        {u}lu\end{tabular}                 & 5.486                   &                     \multicolumn{1}{c|}{0.0037}          & \multicolumn{1}{c|}                         {0.159}            & 0.8123              & 0.9571 & 0.2074 &                        2.4833          \\ \cline{2-9} 
                   %& \begin{tabular}[c]{@{}c@{}}K-Slice: LLaMa; \\ Prompt: SciLitLLM\end{tabular}            & 2.9139                  & \multicolumn{1}{c|}{0.0058}          & \multicolumn{1}{c|}{0.1964}           & 0.823               & 1.2968 & 0.1619 & 2.1000          \\ \cline{2-9} 
                   & \begin{tabular}[c]{@{}c@{}}K-Slice: Mistral; \\ Prompt: T\"{u}lu\end{tabular}               & 6.1125                  & \multicolumn{1}{c|}{0.0029}          & \multicolumn{1}{c|}{0.1465}           & 0.8087              & 1.0791 & \textbf{0.2197}               & 2.4333          \\ \cline{2-9} 
                   %& \begin{tabular}[c]{@{}c@{}}K-Slice: Mistral; \\ Prompt: SciLitLLM\end{tabular}          & 3.3845                  & \multicolumn{1}{c|}{0.0033}          & \multicolumn{1}{c|}{0.2122}           & 0.8206              & 1.3194 & 0.1504 & 2.0167          \\ \cline{2-9}

                   % & \begin{tabular}[c]{@{}c@{}}K-Slice: LLaMa; \\ Prompt: QwQ-32B\end{tabular}          & 5.4111                  & \multicolumn{1}{c|}{0.0019}          & \multicolumn{1}{c|}{0.1545}           & 0.8042              & 1.0791 & 0.1958 & 2.4500          \\ \cline{2-9} 
                   % & \begin{tabular}[c]{@{}c@{}}K-Slice: Mistral; \\ Prompt: QwQ-32B\end{tabular}          & 5.8538                  & \multicolumn{1}{c|}{0.0019}          & \multicolumn{1}{c|}{0.1503}           & 0.8078              & 1.0944 & 0.2066 & 2.4071          \\ \cline{2-9} 
                   % & \begin{tabular}[c]{@{}c@{}}K-Slice: LLaMa; \\ Prompt:\\  DeepSeek-R1-Dist.-Qwen-32B\end{tabular}          & 6.7846                  & \multicolumn{1}{c|}{0.0016}          & \multicolumn{1}{c|}{0.1428}           & 0.8037              & 1.0514 & 0.2087 & 2.2807          \\ \cline{2-9} 
                   % & \begin{tabular}[c]{@{}c@{}}K-Slice: Mistral; \\ Prompt:\\  DeepSeek-R1-Dist.-Qwen-32B\end{tabular}          & 7.2543                  & \multicolumn{1}{c|}{0.0018}          & \multicolumn{1}{c|}{0.1489}           & 0.8092              & 0.8434 & \textbf{0.2255} & 2.2143          \\ \cline{2-9} 
                   & \begin{tabular}[c]{@{}c@{}}K-Slice: LLaMa; \\ Prompt: Sky-T1-32B\end{tabular}          & 6.4486                  & \multicolumn{1}{c|}{0.0020}          & \multicolumn{1}{c|}{0.1761}           & 0.8170              & 1.0804 & 0.2135 & 2.3966          \\ \cline{2-9} 
                   & \begin{tabular}[c]{@{}c@{}}K-Slice: Mistral; \\ Prompt: Sky-T1-32B\end{tabular}          & 7.1965                  & \multicolumn{1}{c|}{0.0022}          & \multicolumn{1}{c|}{0.1933}           & 0.8221              & 1.0948 & 0.2103 & 2.4211          \\ \hline
                   
\multirow{4}{*}{\textsc{TaxoAlign}}                                                                    & \begin{tabular}[c]{@{}c@{}}K-Slice: LLaMa; \\ T- Verbal.:                        T\"{u}lu; \\ T-Refine.: GPT-4o-mini\end{tabular}        &                          1.6687                  & \multicolumn{1}{c|}                                      {\textbf{0.0132}} & \multicolumn{1}{c|}{\textbf{0.2975}}  &                        0.8501              & 1.3244 & 0.1986 & 2.4167 \\ \cline{2-9} 
                   %& \begin{tabular}[c]{@{}c@{}}K-Slice: LLaMa; \\ T- Verbal.: SciLitLLM; \\ T-Refine.: GPT-4o-mini\end{tabular}   & 1.7358 & \multicolumn{1}{c|}{0.0081}          & \multicolumn{1}{c|}{0.29}             & 0.8484              & 1.2956 & 0.1875 & 2.4833          \\ \cline{2-9} 
                   & \begin{tabular}[c]{@{}c@{}}K-Slice: Mistral; \\ T- Verbal.: T\"{u}lu; \\ T-Refine.: GPT-4o-mini\end{tabular}      & 1.668                   & \multicolumn{1}{c|}{0.0051}          & \multicolumn{1}{c|}{0.2974}           & \textbf{0.8517}     & \textbf{1.3635}               & 0.1872 & \textbf{2.5000}                        \\ \hline 
                   %& \begin{tabular}[c]{@{}c@{}}K-Slice: Mistral; \\ T- Verbal.: SciLitLLM; \\ T-Refine.: GPT-4o-mini\end{tabular} & 2.1709                  & \multicolumn{1}{c|}{0.0053}          & \multicolumn{1}{c|}{0.284}            & 0.8484              & 1.265  & 0.1966 & 2.4833          \\ \hline
\end{tabular}
\end{adjustbox}
\caption{Results of our method, \textsc{TaxoAlign}, compared with AutoSurvey, STORM, Topic-only, Topic+Keyphrases and \textsc{TaxoAlign} w/o Taxonomy Verbalization w/o Taxonomy Refinement, on the original test set.}
\label{tab:results}
\end{table*}

\begin{table*}[h]
    \centering
\begin{adjustbox}{width=\linewidth}
\begin{tabular}{c|c|c|ccc|c|c|c}
\hline
\multirow{2}{*}{\textbf{Method}}          & \multirow{2}{*}{\textbf{Model}}   & \multirow{2}{*}{\textbf{$\Delta$}} & \multicolumn{3}{c|}{\textbf{Level-order Traversal}}  & \multirow{2}{*}{\textbf{NSR}} & \multirow{2}{*}{\textbf{NER}} & \multirow{2}{*}{\textbf{\begin{tabular}[c]{@{}c@{}}LLM\\ judge\end{tabular}}} \\ \cline{4-6} & & & \multicolumn{1}{c|}{\textbf{BLEU-2}} & \multicolumn{1}{c|}{\textbf{ROUGE-L}} & \textbf{BERTScore} &        &        &                 \\ \hline

\multirow{2}{*}{\begin{tabular}[c]{@{}c@{}}\textsc{TaxoAlign}\\  w/o Tax. Verbaliz.\\ w/o Tax. Refine. \end{tabular}} & \begin{tabular}[c]{@{}c@{}}K-Slice: LLaMa;\\ Prompt: Tulu\end{tabular} & 6.361 & \multicolumn{1}{c|}{0.0019} & \multicolumn{1}{c|}{0.1643} & 0.8182 & 1.0716 & 0.2683 & 2.275 \\ \cline{2-9}
 %& \begin{tabular}[c]{@{}c@{}}K-Slice: LLaMa;\\ Prompt: SciLitLLM\end{tabular} & 6.5221 & 0.0026 & 0.1957 & 0.8183 & 1.2095 & 0.2267 & 1.9375 \\ \cline{2-9}
 & \begin{tabular}[c]{@{}c@{}}K-Slice: Mistral;\\ Prompt: Tulu\end{tabular} & 7.2083 & \multicolumn{1}{c|}{0.0034} & \multicolumn{1}{c|}{0.1598} & 0.8159 & 1.0737 & 0.2653 & 2.2125 \\ \hline
 %& \begin{tabular}[c]{@{}c@{}}K-Slice: Mistral;\\ Prompt: SciLitLLM\end{tabular} & 5.9387 & 0.0032 & 0.2039 & 0.8211 & \textbf{1.3243} & 0.2176 & 1.875 \\ \hline

\multirow{4}{*}{\textsc{TaxoAlign}} & \begin{tabular}[c]{@{}c@{}}K-Slice: LLaMa; \\ T-Verbal.: Tulu; \\ T-Refine.: gpt-4o-mini\end{tabular} & \textbf{2.1924} & \multicolumn{1}{c|}{0.0058} & \multicolumn{1}{c|}{\textbf{0.3091}}  & \textbf{0.8542}     & \textbf{1.2129} & 0.2566 & 2.2875 \\ \cline{2-9}
 %& \begin{tabular}[c]{@{}c@{}}K-Slice: LLaMa; \\ T-Verbaliz.: SciLitLLM; \\ T-Refine.: gpt-4o-mini\end{tabular}   & 2.5551 & \textbf{0.0127} & 0.3034 & 0.851 & 1.1927 & 0.2614 & 2.2625 \\ \cline{2-9}
 & \begin{tabular}[c]{@{}c@{}}K-Slice: Mistral; \\ T-Verbal.: Tulu; \\ T-Refine.: gpt-4o-mini\end{tabular}      & 2.3617 & \multicolumn{1}{c|}{\textbf{0.013}} & \multicolumn{1}{c|}{0.3004} & 0.8522 & 1.2072 & \textbf{0.2716} & \textbf{2.35} \\ \hline
 %& \begin{tabular}[c]{@{}c@{}}K-Slice: Mistral; \\ T-Verbaliz.: SciLitLLM; \\ T-Refine.: gpt-4o-mini\end{tabular} & 3.1779 & 0.0042 & 0.2845 & 0.8465 & 1.1806 & 0.267 & 2.3125 \\ \hline

\end{tabular}
\end{adjustbox}
\caption{Results of \textsc{TaxoAlign} compared with \textsc{TaxoAlign} w/o Taxonomy Verbalization w/o Taxonomy Refinement on the additional test set of conference papers.}
\label{tab:conf_results}
\end{table*}

\section{Experimental Setup}

\subsection{Base Models}

We use the open-domain \textbf{Mistral-7B-Instruct-v0.3} \cite{jiang2023mistral7b} and \textbf{Meta-Llama-3-8B-Instruct} \cite{grattafiori2024llama3herdmodels} 
%language models 
for extracting the knowledge slices. In the taxonomy verbalization phase, we finetune  \textbf{Llama-3.1-T\"{u}lu-3-8B} \cite{lambert2025tulu3pushingfrontiers} and  \textbf{SciLitLLM1.5-7B} \cite{li2024scilitllmadaptllmsscientific}  respectively. For the refinement stage, we use a closed-domain model \textbf{GPT-4o-mini}. 
For the prompting stage in the \textsc{TaxoAlign} w/o Taxonomy Verbalization w/o Taxonomy Refinement baseline, we also test using three large open-source reasoning models (Qwen’s \textbf{QwQ-32B} \cite{qwq32b}, DeepSeek-AI’s \textbf{DeepSeek-R1-Distill-Qwen-32B} \cite{deepseekai2025deepseekr1incentivizingreasoningcapability} and NovaSky-AI’s \textbf{Sky-T1-32B}\cite{sky_t1_2025}) using the previously generated knowledge slices. Prompts and model details are present in Appendix \ref{appendix:method-prompts}. 
We choose the  T\"{u}lu and the SciLitLLM models for instruction tuning in \textsc{TaxoAlign} as well as prompting because they include extensive scientific research-related data in their pre-training or continual pre-training corpus.

\subsection{Hyperparameter Choices}
In the taxonomy verbalization stage, we instruction-tune LLMs using QLoRA \cite{qlora}. QLoRA uses 4-bit NormalFloat, Double Quantization and Paged Optimizers on the LoRA fine-tuning approach \cite{hu2022lora}.
Each language model is instruction-tuned for $800$ steps with an input context window of $16,384$ and a output context window of $1,024$. The learning rate is $2e-4$ and the training batch size is $1$.
For instruction-tuning, we use simple intuitive prompts based on training data from \textsc{CS-TaxoBench} with Alpaca-like \footnote{\url{ https://huggingface.co/datasets/tatsu-lab/alpaca}}instruction format \cite{alpaca}. The instruction format is given in the Appendix \ref{appendix_subsection:instrction_format}.
We instruct LLMs in our experiments to generate taxonomies with a maximum depth of three. For the taxonomy verbalization part in our method or in any of the baselines, we set $1,024$ as the maximum number of new tokens to be generated by the model. All experiments are done on a single A100.

\section{Results and Analysis}

We evaluated our method and the baseline methods using the proposed metrics. We summarize the results of our experiments in Table \ref{tab:results}. Additional results with more models are in Table \ref{tab:results_appendix} of Appendix \ref{appendix:extended_results}.
We find that the \textsc{TaxoAlign} w/o Taxonomy Verbalization w/o Taxonomy Refinement baseline performs second best to our method 
%in terms of most of the metrics. 
on most metrics. 
This indicates that the knowledge slices are an important tool for this task. This baseline has high average degree score compared to our method that reveals that our Taxonomy Verbalization and Refinement stages could effectively reduce the gap between the generated and gold taxonomy trees while enhancing the quality of node labels. In our experiments, we find that the BLEU-2, ROUGE-L and BERTScore values are much less than what we typically encounter in tasks like machine translation or question answering. This suggests substantial scope for improvement in this task, as a huge gap remains between the human-created and machine-generated results.

\subsection{Structural Similarity}

Our method consistently achieves an average degree score ($\Delta$) close to $1$, while the $\Delta$ value obtained by other baseline methods are much higher. This indicates that our generated tree is much closer to the human-written taxonomy tree in terms of structure. We observe that the $\Delta$ value is closest to $1$ %in the case of  
for 
the Topic-only baseline. This is mainly because when provided with only the topic, the language model generates a small tree due to lack of parametric knowledge, as has been established by the consistent low scores for this baseline on the rest of the metrics. Other baseline methods tend to produce overly large trees ($\Delta > 2.9$) with an excessive number of nodes and branches, increasing the likelihood of hallucinations and structural divergence from the gold-standard tree. We show examples of generated taxonomies using AutoSurvey and STORM in comparison with \textsc{TaxoAlign} in Figures \ref{fig:Taxonomy-autosurvey-example} and \ref{fig:Taxonomy-storm-example} respectively in Appendix \ref{appendix:out_examples}.
%generate huge trees which contain large number of nodes and branches, which increases the chances of hallucinations and demonstrates a marked shift in structure from the gold-standard tree.

\subsection{Semantic Similarity} 

In terms of the level-order traversal, we observe that our method produces comprehensively better BLEU-2, ROUGE-L and BERTScore in all the cases when compared with the baselines. 
%We observe that all the baseline methods have similar kind of scores after level-order traversal, which shows that they generate nodes which have low lexical overlap between the generated and gold data.
We observe that all baselines  have similarly low scores for level-order traversal, indicating that the generated nodes exhibit low lexical overlap with the gold data.
In comparison, our method produces more coherent labels and nodes in the taxonomy tree.

In terms of Node Soft Recall (NSR), our method performs better than the other baselines, showing the similarity between the generated and gold node labels.
The \textsc{TaxoAlign} w/o Taxonomy Verbaliz. w/o Taxonomy Refine. baseline performs better in some cases in terms of the Node Entity Recall metric, which is mainly because this baseline generates large trees, as has been demonstrated by the $\Delta$ value, and larger trees contribute to greater match in the Noun Phrase chunks.
%Using LLM-as-a-judge too, \textsc{TaxoAlign} performs better than the baselines, which reaffirms the results of the above-specified metrics, showing that \textsc{TaxoAlign} generates taxonomies that are closer to the human-written taxonomy in terms of structure and intent. 
Using LLM-as-a-judge, \textsc{TaxoAlign} also outperforms the baselines, reaffirming the metric-based results and showing that it generates taxonomies closer to the human-written ones in both structure and intent.

\subsection{Testing with Conference Survey Papers}

 We tested our three-phase pipeline (that had been trained on 400-instance training data) on the new test set introduced in Section \ref{conf_papers_section}, the results for which are given in Table \ref{tab:conf_results}. Extended results with more models are present in Table \ref{tab:conf_results_appendix} of Appendix \ref{appendix:extended_results}. We observe that our method comfortably outperforms the baseline on this test set too, which is consistent with %our original 
 the results reported in Table \ref{tab:results}.

\subsection{Error Analysis}

%We present an error analysis based on a manual evaluation of different instances from our original test set on our proposed \textsc{TaxoAlign} pipeline:\textcolor{red}{(DKS: Can we add examples of each type of error, maybe in the Appendix?)}\\
We present an error analysis based on a manual evaluation of instances from the %original 
test set in Table \ref{tab:stats} using our proposed \textsc{TaxoAlign} pipeline. For illustration, we provide a example for each of the three stages of the pipeline in Figure \ref{fig:Taxonomy-error-analysis} of Appendix \ref{appendix:error_analysis}. Below, we summarize the common errors observed at each stage of the pipeline. 

\textbf{Knowledge Slices + Prompting:}
Direct generation from knowledge slices creates more verbose taxonomies that contain irrelevant information leading to the  nodes not being very specific or not pertaining to the topic directly. Another major factor is the presence of repeated numbers of the same nodes or sub-trees in the taxonomy.

\textbf{Knowledge Slices + Taxonomy Verbalization:}
Structurally, the taxonomies are closer to the gold standard taxonomies, but there are some factual errors that persist. We observe that the generated taxonomies suffer from the problem of hallucinated node labels or are too short.

\textbf{Knowledge Slices + Taxonomy Verbalization + Taxonomy Refinement:}
The generated trees are more aligned to the gold standard trees in terms of structure and semantic coherence. Still, the generated trees suffer from a low number of layer-wise exact matches. The generated trees are certainly more interpretable than the previous stages and the overall tree also presents a coherent structure.

\section{Human Evaluation}

\begin{table}[t]
    \centering
    \begin{tabular}{c|c|c}
        \hline
      Method & Structure   & Content \\ \hline
      \textsc{TaxoAlign} & $3.17$ & $2.62$ \\
      AutoSurvey & $2.17$ & $2.25$ \\ \hline
    \end{tabular}
    \caption{Mean ratings from human evaluation of the structure and content similarity for \textsc{TaxoAlign} and AutoSurvey.}
    \label{tab:human_eval}
\end{table}

%We use a human-based evaluation to complement the automated evaluation framework. We ask three annotators who posses enough domain knowledge to rate the generated surveys from \textsc{TaxoAlign} and AutoSurvey. We ask the human annotators to assess 
We use human evaluation to complement the automated framework. Three annotators with  domain knowledge were asked to rate the surveys generated by \textsc{TaxoAlign} and AutoSurvey. The annotators are instructed to assess based on (1) the structural commonalities between the gold and generated taxonomy trees and then (2) the semantic coherence of the generated tree with respect to the gold tree. The evaluation is done using a 5-point Likert scale on $20$ randomly sampled data instances from the test set of \textsc{CS-TaxoBench}. The inter-annotator agreement is calculated as $0.61$ and $0.73$ (Krippendorff's $\alpha$). The results are shown in Table \ref{tab:human_eval}. The mean ratings show \textsc{TaxoAlign} outperforms \textsc{AutoSurvey} %both in terms of structural and content quality.
in both structural and content quality.

%To verify the consistency between our LLM-as-a-judge evaluation and human evaluation, we first average the scores given by human evaluators for each taxonomy tree. We then compare these scores with those generated by the LLM using Spearman’s rank correlation coefficient.
To verify the consistency between our LLM-as-a-judge evaluation and the human evaluation, we first average the scores assigned by human annotators for each taxonomy tree. We then compare these with the LLM-generated scores using Spearman’s rank correlation coefficient. 
Thereby, we obtain a Spearman’s rho value of $0.527$ which indicates a strong positive correlation. These results suggest that our LLM-as-a-judge evaluation method aligns well with human preferences, providing a reliable proxy for human judgment.

\section{Conclusion}

%Automation of scholarly taxonomy generation can enable researchers and practitioners to quickly navigate the vast literature. To facilitate this, we present \textsc{CS-TaxoBench}, a benchmark of $460$ taxonomies, and \textsc{TaxoAlign}, a method which uses instruction tuning and refinement on topic-related information extracted from papers. We propose two new metrics and also use previously proposed metrics to evaluate our method with several baselines. It can be observed that our method outperforms the tested baselines on most of the metrics.
Automating scholarly taxonomy generation can help researchers and practitioners efficiently navigate the vast body of scholarly literature. To facilitate this, we present \textsc{CS-TaxoBench}, %a benchmark of $460$ and $80$ taxonomies from journey surveys and  conference surveys, respectively, %$80$ taxonomies from conference surveys, 
a benchmark comprising 460 taxonomies from journey surveys and 80 from conference surveys, 
along with   \textsc{TaxoAlign}, a method that uses instruction tuning and refinement on topic-related information extracted from papers. 
We introduce two new metrics and show that \textsc{TaxoAlign} outperforms the baselines on most evaluation measures.%We introduce two new metrics and use existing ones to evaluate our method against several baselines. Our results show that \textsc{TaxoAlign} outperforms the baselines on most metrics.

%\section*{Acknowledgments}

\section*{Limitations}

We construct \textsc{CS-TaxoBench} from a single journal within a defined time frame to ensure consistency in taxonomy quality. However, additional open-access journals and conference venues could also be explored for future curation.
%We create \textsc{CS-TaxoBench} from a specific journal and consider only a specific time frame for the curation of our dataset. While this is mainly done to keep the quality of the taxonomies consistent, other open-access journals and conference venues could also be explored for future curation. 
%there can be more open-source journals and conference venues from where such taxonomies may be scraped.

We do not focus on the retrieval of the reference papers from a corpus of papers. While this is an important task for end-to-end taxonomy construction given only the taxonomy topic, we focus more on creating and evaluating taxonomies when provided with a set of reference documents.

Although we improve the structure and semantic coherence between the human-written and the generated taxonomies using \textsc{TaxoAlign}, there is a a lot of scope for improvement in this field. Therefore, this is a encouraging field of work in which the community can work in the coming days.

\bibliography{custom}

\appendix

\section{Our Method -- Prompts and Models}
\label{appendix:method-prompts}

\subsection{Models}

\begin{itemize}
    \item \textbf{Mistral-7B-Instruct-v0.3} \cite{jiang2023mistral7b}: The Mistral group of models leverages grouped-query attention (GQA) for faster inference, coupled with sliding window attention (SWA) to effectively handle sequences of arbitrary length with a reduced inference cost. 
    %In comparison to the 0.2 version, this model can effectively process and react to a variety of activities and instructions thanks to its expanded vocabulary of 32,768 and support for v3 Tokenizer. 
    Compared to version 0.2, this model can process and respond more effectively to diverse tasks and instructions, owing to its expanded vocabulary of 32,768 tokens and support for the v3 tokenizer.
    The model can carry out operations that call for outside data since it supports function calling.
    
    \item \textbf{Meta-Llama-3-8B-Instruct} \cite{grattafiori2024llama3herdmodels}: Llama is a family of pre-trained foundational language models that have been open-sourced by Meta in recent times. The Meta-Llama-3-8B-Instruct is trained on a mix of publicly available online data with a knowledge cutoff of March, 2023. The tuned versions of Llama3 use Supervised Fine-Tuning (SFT) and Reinforcement Learning with Human Feedback (RLHF) to align with human preferences.
    
    \item \textbf{Llama-3.1-T\"{u}lu-3-8B} \cite{lambert2025tulu3pushingfrontiers}: T\"{u}lu \cite{wang2023far} is a set of models that are instruction-tuned on LLaMA \cite{touvron2023llamaopenefficientfoundation} using a mixture of publicly available, synthetic and human-created datasets. Building upon the Llama 3.1 basic models, T\"{u}lu-3 \cite{lambert2025tulu3pushingfrontiers} models are trained using Direct Preference Optimization (DPO), Supervised Fine-Tuning (SFT), and a technique called Reinforcement Learning with Verifiable Rewards (RLVR). 
    
    \item \textbf{SciLitLLM1.5-7B} \cite{li2024scilitllmadaptllmsscientific}: It is a very recently released LLM designed for the task of scientific literature understanding that has been trained using both Continual Pre-Training (CPT) and Supervised Fine-Tuning (SFT).This strategy is used on Qwen2.5 to obtain SciLitLLM. The CPT stage uses 73,000 textbooks and 625,000 academic papers, while the SFT stage uses SciLitIns, SciRIFF \cite{wadden2024sciriffresourceenhancelanguage} and Infinity-Instruct\footnote{\url{https://huggingface.co/datasets/BAAI/Infinity-Instruct}}. We use the SciLitLLM 7B\footnote{\url{https://huggingface.co/Uni-SMART/SciLitLLM}} for our experimental purposes.
    
    \item \textbf{QwQ-32B} \cite{qwq32b}: QwQ is designed for complex problem-solving and logical reasoning tasks and is based on Qwen2.5. The model is text-only and focuses on tasks like multi-step reasoning, complex decision-making, and research assistance. 
    
    \item DeepSeek-AI’s \textbf{DeepSeek-R1-Distill-Qwen-32B}  \cite{deepseekai2025deepseekr1incentivizingreasoningcapability}: DeepSeek-R1-Distill-Qwen-32B is an open-source, distilled large language model (LLM) based on the Qwen2.5 32B architecture, utilizing the knowledge from the DeepSeek-R1 reasoning model. It is optimized for language understanding, reasoning, and text generation tasks and is known for outperforming other open-source models, including OpenAI's o1-mini, on various industry benchmarks.
    
    \item \textbf{Sky-T1-32B} \cite{sky_t1_2025}: This model has been developed by the NovaSky team at UC Berkeley. It excels in mathematical and coding reasoning, outperforming some advanced closed-source models and other open-source alternatives on various benchmarks. The model was created by fine-tuning the Qwen 2.5 32B instruct model with a high-quality, 17,000-item dataset.
\end{itemize}

\subsection{Knowledge Slice-Prompt}

\begin{boxD}
You will receive a document and a topic. Your task is to identify the knowledge-slices within the document that are very relevant to the given topic. A knowledge-slice is a piece of information representing the highlights of the document related to the given topic i.e. each knowledge-slice should be such that it both represents an important point in the document, but at the same time, the knowledge-slice should pertain closely to the given topic. Also, the knowledge-slice should not represent any additional information that is not present in the document.

[Document]

document-text

[Topic]

taxonomy-topic

Please ONLY return the relevant knowledge-slices in the form of a list enclosed within square brackets. Your response should be in the following format:

[Knowledge-Slices]

[Knowledge-Slice 1, Knowledge-Slice 2,..., Knowledge-Slice n]

[Your response]

\end{boxD}

\subsection{Taxonomy Verbalization-Prompt}

\begin{boxD}
    A taxonomy is a tree-structured semantic hierarchy that establishes a classification of the existing literature under a common topic. You will receive a taxonomy topic along with a collection of documents. Your task is to create a taxonomy tree using the given topic and based on the highlights of the documents i.e. create new child nodes by identifying generalizable sub-level topics from the document highlights that can act as child nodes to the taxonomy topic, which acts as the root node . The taxonomy tree should be created such that it looks as if all the given documents are a part of the taxonomy. There may be several levels in the tree i.e. each node may contain child nodes, but the total depth of the tree should not exceed three. The topics in all the levels of the tree except the last level must not be too specific so that it can accommodate future sub-topics i.e. child nodes.

    - The nodes at the last level of the hierarchy i.e. the leaf nodes should reflect a single topic instead of a combination of topics.
    
    - Each node label is a small and concise phrase.

[Response Format Instructions]

    - The output tree is to be formatted as shown in the example such that the root node is the taxonomy topic and each child node is connected to its parent.

[Example Output]

example-output     

[Taxonomy Topic]

taxonomy-topic

[Documents]

Doc-1

Doc-2

Doc-3

Please ONLY return the taxonomy tree in the output format as shown in the example above.

[Your response]

\end{boxD}

\subsection{Taxonomy Refinement- Prompt}

\begin{boxD}
    A taxonomy is a tree-structured semantic hierarchy that establishes a classification of the existing literature under a common topic. You will receive a taxonomy tree along with a collection of documents. The root node of the taxonomy tree is the overall taxonomy topic. Your task is to refine the taxonomy tree such that there is a clear connection between the parent node and the subsequent child nodes. Each node must be a well-defined topic that is grounded in the input document highlights. Do not alter the root node of the tree i.e. the taxonomy topic. Your task is to alter the other nodes only if deemed necessary i.e. only if a better viable replacement is found. Please try to adhere to the structure of the given taxonomy tree as much as possible. Only if the given taxonomy tree is restricted to less than five nodes, then generate the taxonomy tree on your own. Strictly adhere to the format of the tree shown here.

[Example Output]

example-output     

[Taxonomy Topic]

taxonomy-topic

[Documents]

Doc-1

Doc-2

Doc-3

Please ONLY return the edited taxonomy tree in the output format as shown in the example above. 

[Your response]

\end{boxD}

\section{LLM-as-a-Judge Prompt}
\label{appendix:llm-eval}

\begin{boxD}
A taxonomy is a tree-structured semantic hierarchy that establishes a classification of the existing literature under a common topic. You are given a gold standard taxonomy tree and a generated taxonomy tree and your task is to respond with an appropriate score after comparing the two. Two taxonomy trees are said to be structurally similar if the number of nodes and branches are similar in number. If one tree has too many or too less nodes and branches than the gold tree, then they are said to be structurally dissimilar. Two taxonomy trees are said to be semantically similar if their nodes have values with close meanings or are matching entirely. Please respond with only the score based on the following criteria:

Score 1: The generated taxonomy has no similarity at all with the gold standard taxonomy i.e. the structure and the intent of the generated taxonomy is totally different from that of the gold standard taxonomy.

Score 2: The generated taxonomy have only a few nodes that has a semantic match with the nodes in the gold standard taxonomy and the structure of the generated taxonomy is a little similar to that of the gold standard taxonomy. The structure of the generated tree is very less similar to the gold standard tree but the intent of both taxonomies is similar. 

Score 3: The generated taxonomy has a reasonable similarity to the generated taxonomy in terms of structural similarity and semantic similarity. The structure of both trees are similar but some nodes are different in the two taxonomies. 

Score 4: The generated taxonomy has good logical consistency with that of the gold standard taxonomy in terms of semantic matching of the nodes between the two with the structure of the generated taxonomy is very similar to that of the gold standard taxonomy. The two taxonomies only differ for a small number of instances.

Score 5: The generated taxonomy is fully similar in terms of semantic matching and structure to the gold standard taxonomy.

Gold Standard Taxonomy:

gold-taxonomy

Generated Taxonomy:

generated-taxonomy

[Your Response]

\end{boxD}

\subsection{Instruction Format for Finetuning}
\label{appendix_subsection:instrction_format}

\begin{boxD}

Below is an instruction that describes a task, paired with an input that provides further context. Write a response that appropriately completes the request.

\#\#\# Instruction:

[Instruction prompt (present in Appendix A.2)]

\#\#\# Input:

[Knowledge slices]

\#\#\# Response:

[Gold Standard Taxonomy Tree]

\end{boxD}

\section{Extended Results}
\label{appendix:extended_results}

We show additional results using a expanded set of models on the original test set and the additional conference paper test in Tables \ref{tab:results_appendix} and \ref{tab:conf_results_appendix} respectively.

\begin{table*}[!htbp]
    \centering
\begin{adjustbox}{width=\linewidth}
\begin{tabular}{c|c|c|ccc|c|c|c}
\hline
\multirow{2}{*}{\textbf{Method}}          & \multirow{2}{*}{\textbf{Model}}   & \multirow{2}{*}{\textbf{$\Delta$}} & \multicolumn{3}{c|}{\textbf{Level-order Traversal}}  & \multirow{2}{*}{\textbf{NSR}} & \multirow{2}{*}{\textbf{NER}} & \multirow{2}{*}{\textbf{\begin{tabular}[c]{@{}c@{}}LLM\\ judge\end{tabular}}} \\ \cline{4-6} & & & \multicolumn{1}{c|}{\textbf{BLEU-2}} & \multicolumn{1}{c|}{\textbf{ROUGE-L}} & \textbf{BERTScore} &        &        &                 \\ \hline

AutoSurvey & Prompt: GPT-4o-mini & 4.4659 & \multicolumn{1}{c|}{0.0016} & \multicolumn{1}{c|}{0.1784} & 0.8256 & 1.0903 & 0.1982 & 2.4333 \\ \hline

STORM & Prompt: GPT-4o-mini & 6.151 & \multicolumn{1}{c|}{0.0012} & \multicolumn{1}{c|}{0.1349} & 0.8166 & 1.0727 & 0.1539 & 2.2000 \\ \hline

Topic only & Prompt: T\"{u}lu & \textbf{1.4274} & \multicolumn{1}{c|}{0.0052} & \multicolumn{1}{c|}{0.2359} & 0.8376 & 1.4187 & 0.1373 & 2.0833 \\ \hline

\multirow{6}{*}{\begin{tabular}[c]{@{}c@{}}Topic\\  + \\ Keyphrases\end{tabular}} & \begin{tabular}[c]{@{}c@{}}Keyphrase: LLaMa; \\ Prompt: T\"{u}lu\end{tabular} & 4.4517 & \multicolumn{1}{c|}{0.0018}          & \multicolumn{1}{c|}{0.1584} & 0.8134 & 1.1103 & 0.1491 & 2.4167 \\ \cline{2-9} 
                   & \begin{tabular}[c]{@{}c@{}}Keyphrase: LLaMa; \\ Prompt: SciLitLLM\end{tabular}          & 8.0766                  & \multicolumn{1}{c|}{0.0022}          & \multicolumn{1}{c|}{0.192}            & 0.8168              & 1.2170 & 0.1578 & 1.6833          \\ \cline{2-9} 
                   & \begin{tabular}[c]{@{}c@{}}Keyphrase: Mistral; \\ Prompt: T\"{u}lu\end{tabular}             & 4.91                    & \multicolumn{1}{c|}{0.0014}          & \multicolumn{1}{c|}{0.1432}           & 0.8100              & 1.0996 & 0.1640 & 2.4167          \\ \cline{2-9} 
                   & \begin{tabular}[c]{@{}c@{}}Keyphrase: Mistral; \\ Prompt: SciLitLLM\end{tabular}        & 6.6771                  & \multicolumn{1}{c|}{0.0029}          & \multicolumn{1}{c|}{0.1676}           & 0.8084              & 1.2522 & 0.1670 & 1.6500          \\ \hline
\multirow{2}{*}{\begin{tabular}[c]{@{}c@{}}\textsc{TaxoAlign}\\  w/o \\ Taxonomy\\ Verbalization\\ w/o \\Taxonomy \\Refinement \end{tabular}} & \begin{tabular}[c]{@{}c@{}}K-Slice: LLaMa; \\ Prompt: T\"{u}lu\end{tabular}                 & 5.486                   & \multicolumn{1}{c|}{0.0037}          & \multicolumn{1}{c|}{0.159}            & 0.8123              & 0.9571 & 0.2074 & 2.4833          \\ \cline{2-9} 
                   & \begin{tabular}[c]{@{}c@{}}K-Slice: LLaMa; \\ Prompt: SciLitLLM\end{tabular}            & 2.9139                  & \multicolumn{1}{c|}{0.0058}          & \multicolumn{1}{c|}{0.1964}           & 0.823               & 1.2968 & 0.1619 & 2.1000          \\ \cline{2-9} 
                   & \begin{tabular}[c]{@{}c@{}}K-Slice: Mistral; \\ Prompt: T\"{u}lu\end{tabular}               & 6.1125                  & \multicolumn{1}{c|}{0.0029}          & \multicolumn{1}{c|}{0.1465}           & 0.8087              & 1.0791 & \textbf{0.2197}               & 2.4333          \\ \cline{2-9} 
                   & \begin{tabular}[c]{@{}c@{}}K-Slice: Mistral; \\ Prompt: SciLitLLM\end{tabular}          & 3.3845                  & \multicolumn{1}{c|}{0.0033}          & \multicolumn{1}{c|}{0.2122}           & 0.8206              & 1.3194 & 0.1504 & 2.0167          \\ \cline{2-9}

                   & \begin{tabular}[c]{@{}c@{}}K-Slice: LLaMa; \\ Prompt: QwQ-32B\end{tabular}          & 5.4111                  & \multicolumn{1}{c|}{0.0019}          & \multicolumn{1}{c|}{0.1545}           & 0.8042              & 1.0791 & 0.1958 & 2.4500          \\ \cline{2-9} 
                   & \begin{tabular}[c]{@{}c@{}}K-Slice: Mistral; \\ Prompt: QwQ-32B\end{tabular}          & 5.8538                  & \multicolumn{1}{c|}{0.0019}          & \multicolumn{1}{c|}{0.1503}           & 0.8078              & 1.0944 & 0.2066 & 2.4071          \\ \cline{2-9} 
                   & \begin{tabular}[c]{@{}c@{}}K-Slice: LLaMa; \\ Prompt:\\  DeepSeek-R1-Dist.-Qwen-32B\end{tabular}          & 6.7846                  & \multicolumn{1}{c|}{0.0016}          & \multicolumn{1}{c|}{0.1428}           & 0.8037              & 1.0514 & 0.2087 & 2.2807          \\ \cline{2-9} 
                   & \begin{tabular}[c]{@{}c@{}}K-Slice: Mistral; \\ Prompt:\\  DeepSeek-R1-Dist.-Qwen-32B\end{tabular}          & 7.2543                  & \multicolumn{1}{c|}{0.0018}          & \multicolumn{1}{c|}{0.1489}           & 0.8092              & 0.8434 & \textbf{0.2255} & 2.2143          \\ \cline{2-9} 
                   & \begin{tabular}[c]{@{}c@{}}K-Slice: LLaMa; \\ Prompt: Sky-T1-32B\end{tabular}          & 6.4486                  & \multicolumn{1}{c|}{0.0020}          & \multicolumn{1}{c|}{0.1761}           & 0.8170              & 1.0804 & 0.2135 & 2.3966          \\ \cline{2-9} 
                   & \begin{tabular}[c]{@{}c@{}}K-Slice: Mistral; \\ Prompt: Sky-T1-32B\end{tabular}          & 7.1965                  & \multicolumn{1}{c|}{0.0022}          & \multicolumn{1}{c|}{0.1933}           & 0.8221              & 1.0948 & 0.2103 & 2.4211          \\ \hline
                   
\multirow{10}{*}{\textsc{TaxoAlign}}               & \begin{tabular}[c]{@{}c@{}}K-Slice: LLaMa; \\ T- Verbal.: T\"{u}lu; \\ T-Refine.: GPT-4o-mini\end{tabular}        & 1.6687                  & \multicolumn{1}{c|}{\textbf{0.0132}} & \multicolumn{1}{c|}{\textbf{0.2975}}  & 0.8501              & 1.3244 & 0.1986 & 2.4167          \\ \cline{2-9} 
                   & \begin{tabular}[c]{@{}c@{}}K-Slice: LLaMa; \\ T- Verbal.: SciLitLLM; \\ T-Refine.: GPT-4o-mini\end{tabular}   & 1.7358                  & \multicolumn{1}{c|}{0.0081}          & \multicolumn{1}{c|}{0.29}             & 0.8484              & 1.2956 & 0.1875 & 2.4833          \\ \cline{2-9} 
                   & \begin{tabular}[c]{@{}c@{}}K-Slice: Mistral; \\ T- Verbal.: T\"{u}lu; \\ T-Refine.: GPT-4o-mini\end{tabular}      & 1.668                   & \multicolumn{1}{c|}{0.0051}          & \multicolumn{1}{c|}{0.2974}           & \textbf{0.8517}     & \textbf{1.3635}               & 0.1872 & \textbf{2.5000}                        \\ \cline{2-9} 
                   & \begin{tabular}[c]{@{}c@{}}K-Slice: Mistral; \\ T- Verbal.: SciLitLLM; \\ T-Refine.: GPT-4o-mini\end{tabular} & 2.1709                  & \multicolumn{1}{c|}{0.0053}          & \multicolumn{1}{c|}{0.284}            & 0.8484              & 1.265  & 0.1966 & 2.4833          \\ \hline
\end{tabular}
\end{adjustbox}
\caption{Results of our method compared with baselines like AutoSurvey, Topic-only, Topic+Keyphrases and\textsc{TaxoAlign} w/o Taxonomy Verbalization w/o Taxonomy Refinement.}
\label{tab:results_appendix}
\end{table*}

\begin{table*}[t]
    \centering
\begin{adjustbox}{width=\linewidth}
\begin{tabular}{c|c|c|ccc|c|c|c}
\hline
\multirow{2}{*}{\textbf{Method}}          & \multirow{2}{*}{\textbf{Model}}   & \multirow{2}{*}{\textbf{$\Delta$}} & \multicolumn{3}{c|}{\textbf{Level-order Traversal}}  & \multirow{2}{*}{\textbf{NSR}} & \multirow{2}{*}{\textbf{NER}} & \multirow{2}{*}{\textbf{\begin{tabular}[c]{@{}c@{}}LLM\\ judge\end{tabular}}} \\ \cline{4-6} & & & \multicolumn{1}{c|}{\textbf{BLEU-2}} & \multicolumn{1}{c|}{\textbf{ROUGE-L}} & \textbf{BERTScore} &        &        &                 \\ \hline

\multirow{2}{*}{\begin{tabular}[c]{@{}c@{}}\textsc{TaxoAlign}\\  w/o Tax. Verbaliz.\\ w/o Tax. Refine. \end{tabular}} & \begin{tabular}[c]{@{}c@{}}K-Slice: LLaMa;\\ Prompt: Tulu\end{tabular} & 6.361 & \multicolumn{1}{c|}{0.0019} & \multicolumn{1}{c|}{0.1643} & 0.8182 & 1.0716 & 0.2683 & 2.275 \\ \cline{2-9}
 & \begin{tabular}[c]{@{}c@{}}K-Slice: LLaMa;\\ Prompt: SciLitLLM\end{tabular} & 6.5221 & \multicolumn{1}{c|}{0.0026} & \multicolumn{1}{c|}{0.1957} & 0.8183 & 1.2095 & 0.2267 & 1.9375 \\ \cline{2-9}
 & \begin{tabular}[c]{@{}c@{}}K-Slice: Mistral;\\ Prompt: Tulu\end{tabular} & 7.2083 & \multicolumn{1}{c|}{0.0034} & \multicolumn{1}{c|}{0.1598} & 0.8159 & 1.0737 & 0.2653 & 2.2125 \\ \cline{2-9}
 & \begin{tabular}[c]{@{}c@{}}K-Slice: Mistral;\\ Prompt: SciLitLLM\end{tabular} & 5.9387 & \multicolumn{1}{c|}{0.0032} & \multicolumn{1}{c|}{0.2039} & 0.8211 & \textbf{1.3243} & 0.2176 & 1.875 \\ \hline

\multirow{10}{*}{\textsc{TaxoAlign}} & \begin{tabular}[c]{@{}c@{}}K-Slice: LLaMa; \\ T-Verbaliz.: Tulu; \\ T-Refine.: gpt-4o-mini\end{tabular} & \textbf{2.1924} & \multicolumn{1}{c|}{0.0058} & \multicolumn{1}{c|}{\textbf{0.3091}}  & \textbf{0.8542}     & 1.2129 & 0.2566 & 2.2875 \\ \cline{2-9}
 & \begin{tabular}[c]{@{}c@{}}K-Slice: LLaMa; \\ T-Verbaliz.: SciLitLLM; \\ T-Refine.: gpt-4o-mini\end{tabular}   & 2.5551 & \multicolumn{1}{c|}{\textbf{0.0127}} & \multicolumn{1}{c|}{0.3034} & 0.851 & 1.1927 & 0.2614 & 2.2625 \\ \cline{2-9}
 & \begin{tabular}[c]{@{}c@{}}K-Slice: Mistral; \\ T-Verbaliz.: Tulu; \\ T-Refine.: gpt-4o-mini\end{tabular}      & 2.3617 & \multicolumn{1}{c|}{0.013} & \multicolumn{1}{c|}{0.3004} & 0.8522 & 1.2072 & \textbf{0.2716} & \textbf{2.35} \\ \cline{2-9}
 & \begin{tabular}[c]{@{}c@{}}K-Slice: Mistral; \\ T-Verbaliz.: SciLitLLM; \\ T-Refine.: gpt-4o-mini\end{tabular} & 3.1779 & \multicolumn{1}{c|}{0.0042} & \multicolumn{1}{c|}{0.2845} & 0.8465 & 1.1806 & 0.267 & 2.3125 \\ \hline

\end{tabular}
\end{adjustbox}
\caption{Results of \textsc{TaxoAlign} compared with \textsc{TaxoAlign} w/o Taxonomy Verbalization w/o Taxonomy Refinement.}
\label{tab:conf_results_appendix}
\end{table*}

\section{Output Example Comparison}
\label{appendix:out_examples}

\begin{figure*}[t]
  \centering
  \includegraphics[width=\linewidth]{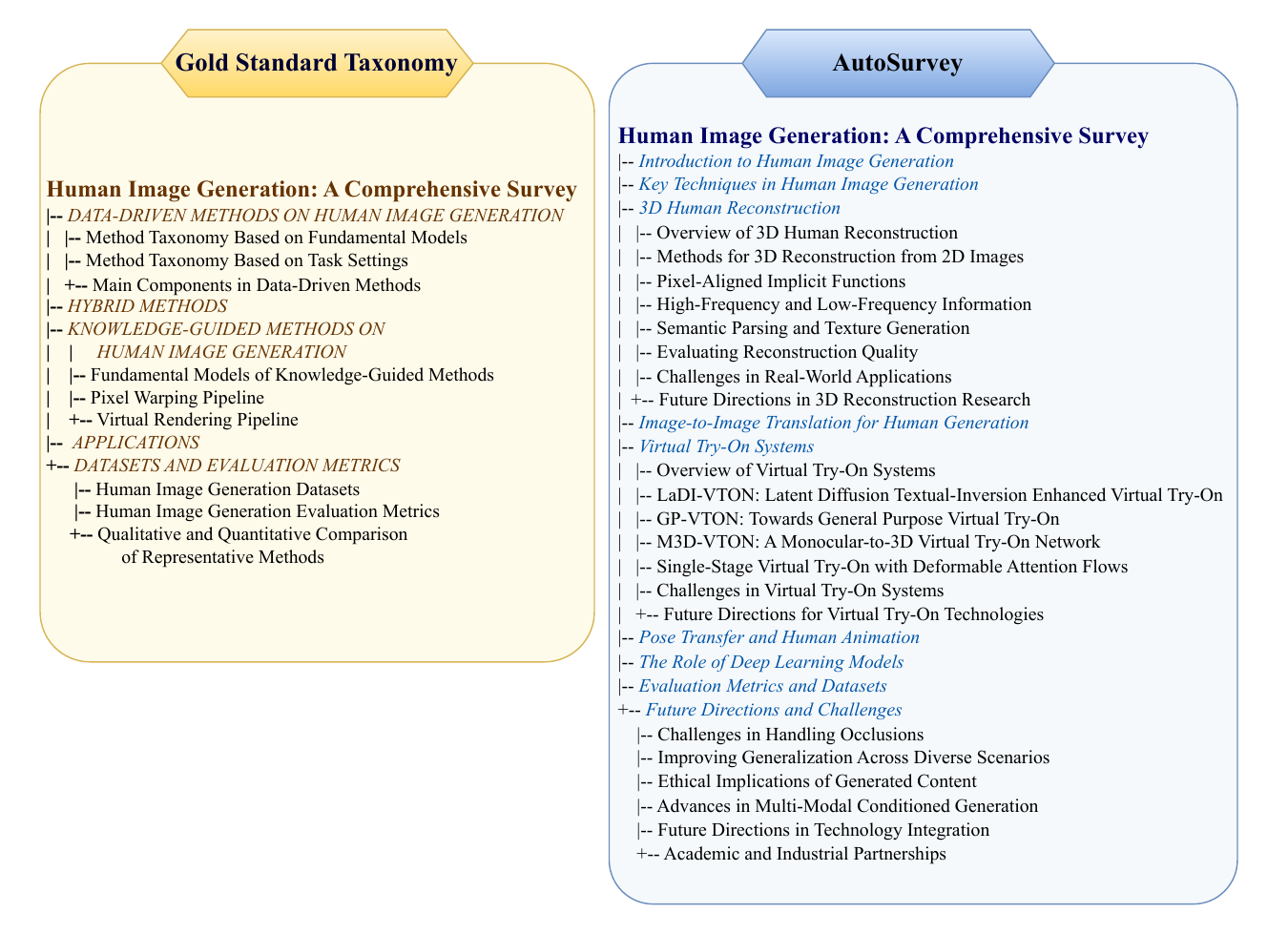}
  \caption{A comparison of a gold standard taxonomy tree and a generated taxonomy tree using AutoSurvey.}
  \label{fig:Taxonomy-autosurvey-example}
\end{figure*}

\begin{figure*}[t]
  \centering
  \includegraphics[width=\linewidth]{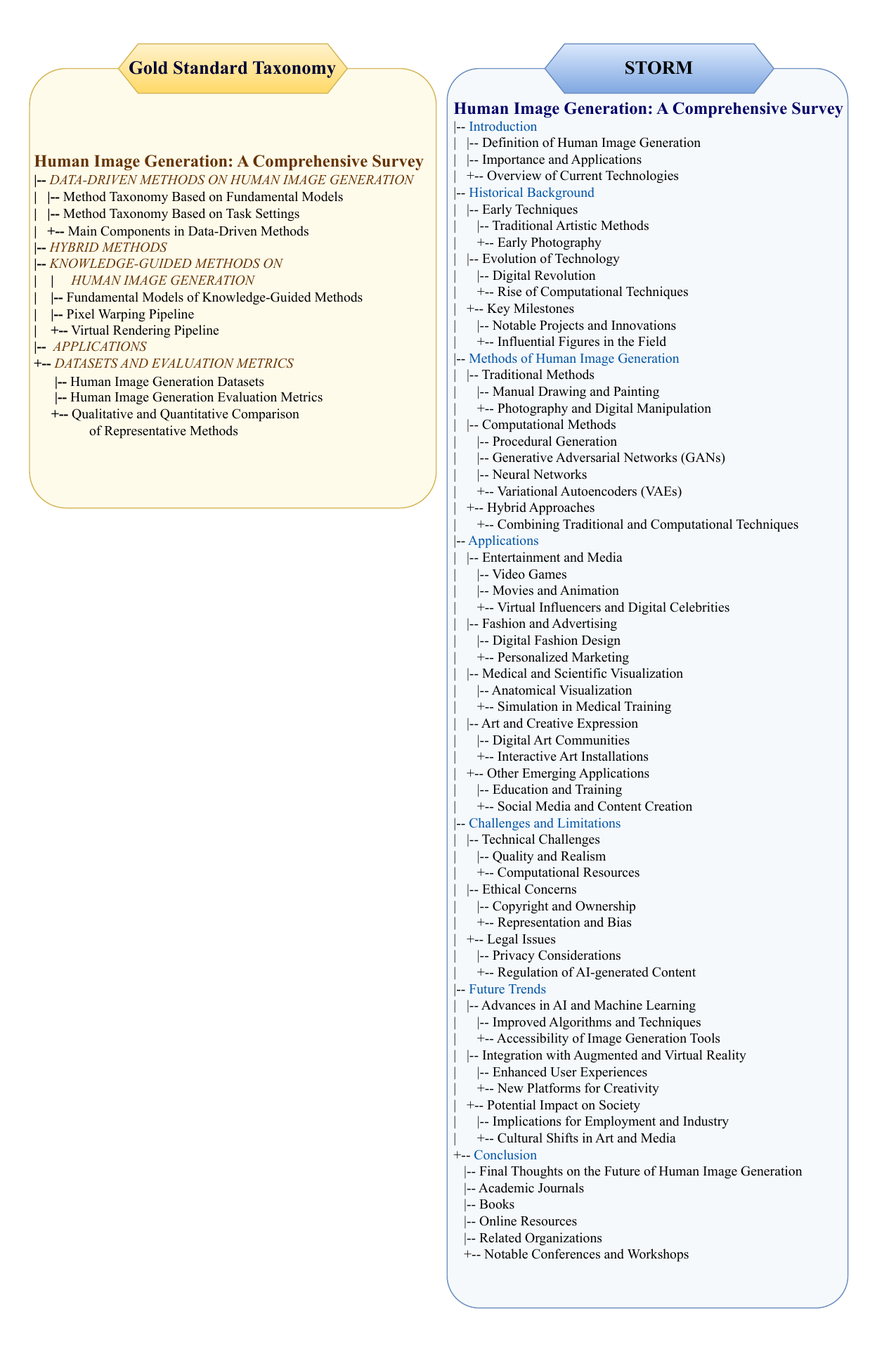}
  \caption{A comparison of a gold standard taxonomy tree and a generated taxonomy tree using STORM.}
  \label{fig:Taxonomy-storm-example}
\end{figure*}

We see in Figure \ref{fig:Taxonomy-autosurvey-example} and \ref{fig:Taxonomy-storm-example} that the taxonomy trees generated using \textsc{TaxoAlign} are much less verbose than the corresponding taxonomy trees generated using AutoSurvey or STORM.

\section{Error Analysis}
\label{appendix:error_analysis}

\begin{figure*}[h]
  \centering
  \includegraphics[width=\linewidth]{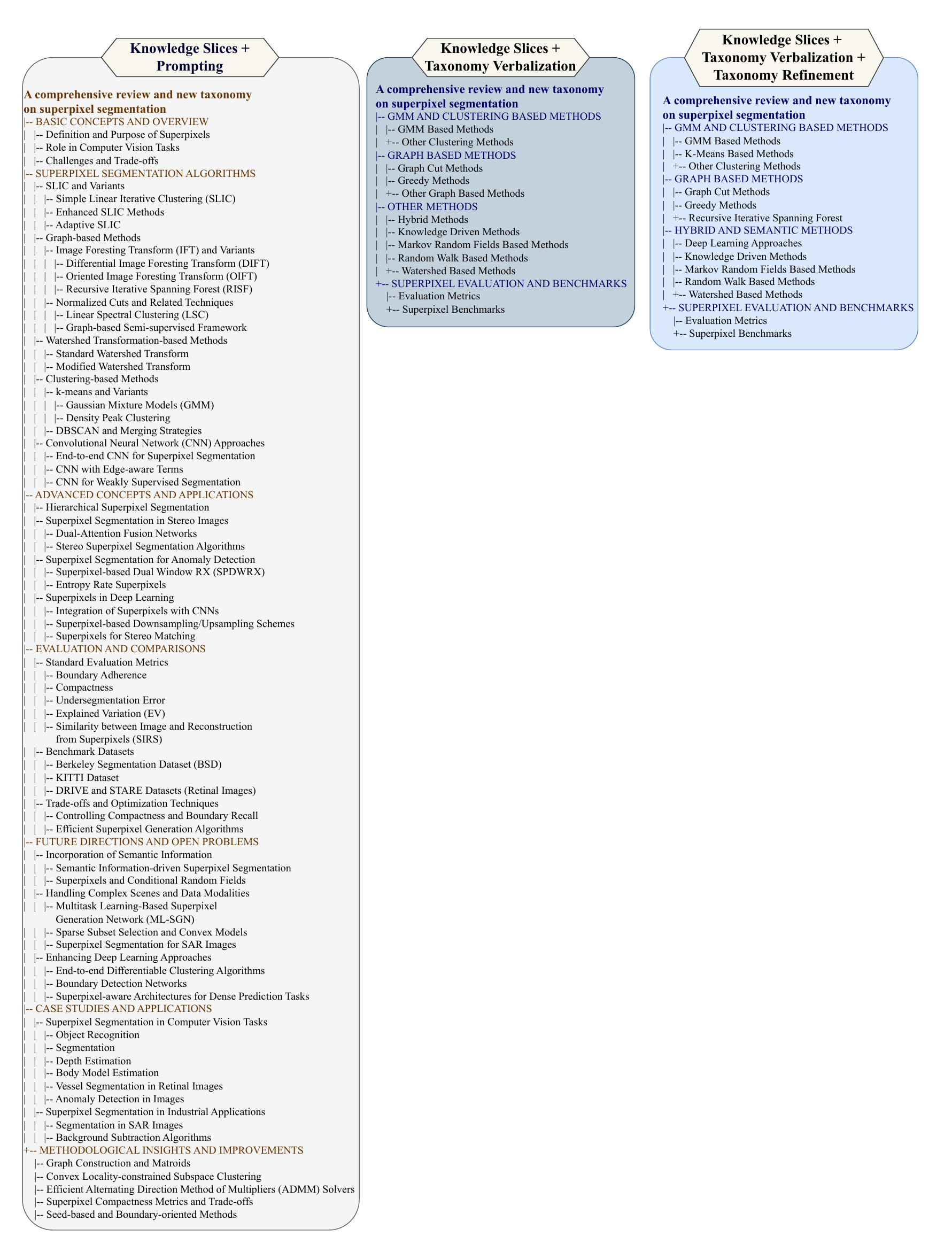}
  %\textcolor{red}{(YH: please make the curve of the box sharper}\\
  \caption{A comparison of the results at the end of each stage of the \textsc{TaxoAlign} pipeline for the topic "A comprehensive review and new taxonomy on super-pixel segmentation". The generated taxonomy shown here uses Mistral-7B-Instruct-v0.3 for creation of knowledge slices and Llama-3.1-T\"{u}lu-3-8B for the taxonomy verbalization component and GPT-4o-mini for refining the generated taxonomy.}
  \label{fig:Taxonomy-error-analysis}
\end{figure*}

We show an example of the results obtained in the three stages of our \textsc{TaxoAlign} pipeline in Figure \ref{fig:Taxonomy-error-analysis}. The stages are Knowledge Slices + Prompting, Knowledge Slices + Taxonomy Verbalization and Knowledge Slices + Taxonomy Verbalization + Taxonomy Refinement.

\end{document}